\documentclass[runningheads,a4paper]{llncs}

\usepackage{amssymb}
\usepackage{graphicx}
\usepackage{pgfplots}
\pgfplotsset{compat=1.5}
\usepackage{subfigure}
\usepackage{subfig}
\usepackage{mwe}
\usepackage{algorithm}
\usepackage{algpseudocode}
\usepackage{float}

\usepackage{url}
\urldef{\mailsa}\path|markku.hinkka@aalto.fi|
\urldef{\mailsb}\path|teemu.lehto@qpr.com|    
\urldef{\mailsc}\path|keijo.heljanko@aalto.fi|
\urldef{\mailsd}\path|alex.jung@aalto.fi|
\newcommand{\keywords}[1]{\par\addvspace\baselineskip
	\noindent\keywordname\enspace\ignorespaces#1}

\begin{document}
	
\mainmatter  

\title{Structural Feature Selection for Event Logs}

\titlerunning{Structural Feature Selection for Event Logs}

%
%
\author{Markku Hinkka\inst{1,2} \and Teemu Lehto\inst{1,2} \and Keijo Heljanko\inst{1,3} \and Alexander Jung\inst{1}}
\authorrunning{Markku Hinkka, Teemu Lehto, Keijo Heljanko, Alexander Jung}

\institute{Aalto University, School of Science, Department of Computer Science, Finland
	\and
	QPR Software Plc, Finland \\
	\and
	HIIT Helsinki Institute for Information Technology
	\\
	\mailsa, \mailsb, \mailsc, \mailsd\\
}

%
%

\toctitle{Structural Feature Selection for Event Logs}
\tocauthor{Markku Hinkka, Teemu Lehto, Keijo Heljanko, Alexander Jung}
\maketitle

\begin{abstract}
We consider the problem of classifying business process instances based on structural features derived from event logs. The main motivation is to provide machine learning based techniques with quick response times for interactive computer assisted root cause analysis. In particular, we create structural features from process mining such as activity and transition occurrence counts, and ordering of activities to be evaluated as potential features for classification. We show that adding such structural features increases the amount of information thus potentially increasing classification accuracy. However, there is an inherent trade-off as using too many features leads to too long run-times for machine learning classification models. One way to improve the machine learning algorithms' run-time is to only select a small number of features by a feature selection algorithm. However, the run-time required by the feature selection algorithm must also be taken into account. Also, the classification accuracy should not suffer too much from the feature selection. The main contributions of this paper are as follows: First, we propose and compare nine different feature selection algorithms by means of an experimental setup comparing their classification accuracy and achievable response times. Second, we discuss the potential use of feature selection results for computer assisted root cause analysis as well as the properties of different types of structural features in the context of feature selection.

\end{abstract}

\section{Introduction}
\label{introduction}

In Process Mining, unstructured \textit{event logs} generated by systems in business processes are used to automatically build real-life process definitions and as-is models behind those event logs. There is a growing need to be able to predict properties of newly added event log cases, or process instances, based on case data imported earlier into the system. In order to be able to predict properties of the new cases, as much information as possible should be collected that is related to the event log cases and relevant to the properties to be predicted. Based on this information, a model of the system creating the event logs can be created. In our approach, the model creation is performed using machine learning techniques.

One good source of additional case related features is the information stored into the \textit{sequence of activities} visited by cases. This information includes, e.g., number of times an event log case has visited a certain \textit{activity} and the number of times a case has \textit{transitioned} between two specific activities. Features collected in this fashion are often highly dependent on each other. E.g., a patient whose visit to hospital takes long time (outcome) has quite often been in surgery from which he/she has moved into a ward. In this example, we already can easily find three structural features of which any can be used to predict whether the visit lasted long or not: visited surgery, transitioned from surgery to ward, visited ward. However, depending on all the other patients in the data set, it may be that there are no cases where only a subset of these three features occurs, thus making it redundant to have all three features taken into account when building a model for prediction purposes. Thus, one feature could well be enough to give as accurate predictions as having them all.

Another important aspect in Process Mining is that it is often desired to be able to show dependencies between features. Thus, selecting a feature selection algorithm that produces also this information for minimal extra cost is often tempting. For this purpose, the list of the most relevant features and the extent of their contribution should somehow be returned. One example of this kind of root cause analysis technique is influence analysis described in~\cite{DBLP:conf/bpm/LehtoHH16}. 

The primary motivation for this paper is the need to perform classification based on structural features originating from \textit{activity sequences} in event logs as accurately as possible and using a minimum amount of computing resources and maximizing the throughput in order to be able to use the method even in some interactive scenarios. This motivation comes from the need to build a system that can do classification and root cause analysis activities accurately on user configurable phenomena based on huge event logs collected and analyzed, e.g., using \textit{Big Data processing frameworks} and methods such as those discussed in our earlier paper~\cite{hinkka2016assessing}. The response time of this classification system should be good enough to be used as part of a web browser based interactive process mining tool where user wants to perform classifications and expects classification results to be shown within a couple of seconds.

As an extension to the original paper presented in BPAI 2017 conference, we have increased the amount of feature selection methods by examining 3 new feature selection methods and compared their performance with the original set of algorithms. The first new tested algorithm is Fisher scoring based approach, whereas the two other algorithms are hybrid algorithms mixing Clustering separately with mRMR and Fisher scoring algorithms. Some additional details were also added into the discussion of relevant concepts including a more detailed study on related works by adding a new Section~\ref{complexstructuralfeatures} discussing structural features that are conceptually and also computationally more complex structural features often referred to in related literature. In the experiment sections we included three new publicly available BPI Challenge data sets to be used in the tests.

The rest of this paper is structured as follows: Section~\ref{problemsetup} introduces main concepts related to this paper. Section~\ref{featureselectionmethods} discusses the feature selection concept and gives brief introduction to the methods used in this paper. Section~\ref{testsetup} will then present a framework used for comparing performance of the selected feature selection approaches. The results of the tests will be presented in Section~\ref{testresults} after which we will discuss related work in Section~\ref{relatedwork}. Finally, Section~\ref{conclusions} draws the conclusions from the test results.

\section{Problem Setup}
\label{problemsetup}

The concepts and terminology used throughout this paper mostly follow those commonly used in process mining and machine learning
communities. However, the following subsections will provide some some short introduction to the most important concepts related to this paper.

\subsection{Concepts}
\label{concepts}

An \textit{event log} can be any ordered list of records known as \textit{events}. Every event has at least a \textit{case identifier}, an \textit{activity identifier} and some additional property such as a time stamp that can be used to put the events into some deterministic order. A \textit{case identifier} is used to group \textit{events} belonging somehow into some common contexts. This could be, e.g., session identifier in a web server log, a customer or order identifier in a customer and order handling system such as CRM. Thus, for every event, an unambiguous \textit{case} can be identified which represents a collection of \textit{events} belonging to the same process. The events for a case are represented in the form of a \textit{trace}, i.e., a sequence of unique events~\cite{van2011process}. In addition to the properties listed above, every event can also include any number of additional \textit{event attributes}. Similarly, every case can include any number of additional \textit{case attributes}. A \textit{transition} represents the transition between two successive activities within a case.

In this paper, the term \textit{activity sequence} is used to represent the ordered set of activity labels within cases. \textit{Activity sequences} are represented as strings using a notation used by Aalst et al.~\cite{van2011process} augmented by "virtual" start and end \textit{activities} making it possible to deal with case start and end events and transitions leading to them from outside of the process using similar notation as for normal activity to activity transitions. $S$ is used to denote this virtual \textit{start event} and $E$ is used for \textit{end event}.

E.g., a activity pattern representation $\langle S,a,b,c \rangle$ represents all the \textit{activity sequences} which start from \textit{activity} labeled \textbf{a}, since there is \textbf{S} -activity in the beginning of the representation. \textbf{a} is then followed by \textbf{b} and \textbf{c} activities. Thus, this pattern can match both of the following activity sequences: 
$\langle S,a,b,c,E \rangle$, $\langle S,a,b,c,d,E \rangle$.

\subsection{Structural Features}
\label{structuralfeatures}

As opposed to normal case attributes added to cases in event logs, structural feature term in this paper is used for representing properties of activity sequences of cases. Thus, they can be derived directly from actual activity sequences without need to include any additional custom properties. Having a case identifier, activity identifier and order information such as a time stamp for each event occurrence, is enough. In order to simplify the tests and keep requirements for applying the results of this paper to its minimum, we decided to only concentrate on structural features as predictors in this paper. However, in real use cases, the best results are achieved by including also all the available additional case attributes such as duration, age, etc. into features from which the feature selection is performed~\cite{DBLP:conf/bpm/LeontjevaCFDM15}.

There are several different types of structural features to select from. In this paper, we use notations similar to those used in regular expressions~\cite{thompson1968programming} combined with notation commonly used for activity sequences~\cite{van2011process}. The patterns we focused in this paper are listed in Table~\ref{table:examplestructuralfeatures} with examples of matches when the sequence of activities is illustrated as \textbf{$\langle S, a, b, b, E \rangle$}. 

\begin{table}[!t]
	\centering
	\begin{tabular}{lll}
		\hline
		Pattern / predictor type & Example sequences(s) \\ \hline
		Activity & $\langle a \rangle, \langle b \rangle$ \\
		Transition / 2-grams & $\langle S, a \rangle, \langle a, b \rangle, \langle b, b \rangle, \langle b, E \rangle$ \\
		Starter & $\langle S, a \rangle$ \\
		Finisher & $\langle b, E \rangle$ \\
		Ordering & $\langle a \rangle \rightarrow \langle b \rangle$ \\
	\end{tabular}
	\caption{Structural Feature Types}
	\label{table:examplestructuralfeatures}
\end{table}

For every predictor type listed in the Table~\ref{table:examplestructuralfeatures}, there can be several possible implementations. In this paper, we consider structural features of \textit{activity} and \textit{2-gram} predictor types to be such that their values correspond to the number of occurrences of that pattern within each activity sequence. \textit{Starter and finisher} predictor types however are boolean values indicating whether that pattern is valid for an activity sequence. Order feature type is considered to be a boolean value such that it is true only if the first occurrences of both ends of the order relation are in the specified order. 

The difference between 2-gram and order pattern is that order allows any number of activities to be between the activities of the ordering relation, whereas in 2-gram, the activities of the relation must be successive in the whole sequence of activities. The importance of predictor types also depends very heavily on the type of the data set and the scenario being predicted. 

One more factor to take into account when selecting the actual features is how to handle situations where a feature has more than two different values. E.g., a patient may have visited surgery multiple times while visiting a hospital. In some cases, depending on what we are trying to predict and what kind of prediction models are to be built, it could be better to split these kinds of features into several boolean features. Thus, e.g., we could have a feature for a patient having visited surgery 4 times. However, one has to be careful when to split features like this in order to avoid an explosion in the number of features created for each original feature.

One final step before passing features to the actual model training is to identify and remove any duplicate features that have behaved identically through the whole training set. Some training methods do this automatically, but some do not. 

\subsection{More Complex Structural Features}
\label{complexstructuralfeatures}

As we were interested in minimizing the response time, we decided to consider only the patterns listed in the previous chapter because having more complex patterns would have made the combinations of different predictor types and features to become too big to be able to perform exhaustive tests for and leading to the problem of \textit{curse of dimensionality}~\cite{DBLP:books/mk/HanK2000}. Also, the extraction of all the feature types presented in the literature would have required much more computation time than the selected relatively simple features used in this work. However, in this chapter we briefly explore some of the other widely used feature types.

Some of the most studied patterns in the process mining community have been related to discovering repeats such as tandem- and maximal repeats ~\cite{DBLP:conf/bpm/BoseA09}\cite{bose2009trace}\cite{DBLP:conf/otm/NguyenDRMS14} within the sequences of activities and discriminative patterns ~\cite{DBLP:conf/otm/NguyenDRMS14}\cite{lo2009classification}\cite{cheng2007discriminative}.

For example, maximal repeat $\alpha$ is defined as being a subsequence of a sequence of activities $S$ where $\alpha$ occurs at least twice within $S$  ~\cite{bose2009trace} and extending $\alpha$ to include an additional character to left or right of the sequence would break the inequality. Similarly, a tandem repeat $\alpha$ is defined as being a subsequence of a sequence of activities $S$ where $\alpha$ occurs successively multiple times $S$. Both of these repeat types can be extracted efficiently using suffix trees~\cite{gusfield1997algorithms}. However, due to relative complexity of feature extraction from the source event log, we decided not to explore these feature types in this paper.

Discriminative patterns are referred by Lo et. al. in their article\cite{lo2009classification} are capturing repetitive iterative patterns within the sequences of activities in traces. After this, a feature selection algorithm based on Fisher scoring~\cite{DBLP:books/lib/DudaHS01} is used to select the most discriminating features to be used for actual classification. This method picks the most important repetitive iterative patterns and uses only them as the set of features. The biggest difference this algorithm has to the methods we are using in this paper is that they concentrated only on iterative patterns and that they use only Fisher scoring -based algorithm. Due to the complexities involved in extracting iterative patterns, we decided to include in our paper just the portion where Fisher scoring is used for feature selection. 

One solution for overcoming the problem of selecting the features in classification is also to use a feature selection technique that does not require specifying the value of each feature separately as input to the model training. One such solution is to use neural networks and deep learning based techniques such as Long short-term memory~\cite{hochreiter1997long}\cite{DBLP:journals/corr/EvermannRF16}. The problem with this approach for our purposes is that the resulting model can not be easily used for root cause analysis and to find out what exact structural features may be causing the observed effect.

\subsection{Classification}
\label{classification}

Since all the tests performed in this paper are performed on data sets consisting of already completed cases, we are performing \textit{classification} using machine learning prediction algorithms. Classification in machine learning usually consists of two phases: training a model and performing the actual classifications using the trained model. In this paper, we concentrate on building the classification model using \textit{supervised learning methods}, where algorithms are trained using \textit{predictors}, together with their \textit{outcomes}. A core part of the model building is the selection of \textit{features} to be used as predictors. Often the more you have independent features that may have effect in the outcome, the better. As shown in Section~\ref{structuralfeatures}, a lot of features can be created directly from the activity sequences of the cases themselves. 

Another important factor that has direct effect to the prediction performance of the model is the algorithm that is used for building the model and making the predictions. In this paper, we focus on the feature selection part. However, we need to also validate the performance of the feature selection using a set of algorithms. In order to minimize the skew in the results caused by the validation algorithm itself, we decided to compare the efficiency of the selected features using two different approaches. First, for a given set of selected features, we determined the prediction accuracy obtained by a particular supervised learning method, i.e., the gradient boosting machine (GBM). GBM was selected due to its good reputation~\cite{Ogutu2011,DBLP:books/mk/HanK2000} and performance in both accuracy and response times in our own tests. The second method was to approximate the mutual information score~\cite{meyer2008information} between each of the selected set of features and two different data sets. The first data set consisted of all the available predictors without any feature selection. The second data set consisted only of the outcomes to be predicted.

As we are concentrating on features originating from process mining, at the granularity level of a case, the prediction inputs that are usually used in the field are actually custom case properties such as the customer name or an identifier of the owner of the case. The outcomes that we want to predict are usually values of some custom case properties or some calculated case content dependent values such as durations, some kind of cost of the case or some other metrics measuring the quality of the case. In this paper, we concentrate only in features inherent to the activity sequences inside cases and measure how well certain outcomes can be predicted only based on those features.

The used data set is split into two parts: training and test. Training data set is used for two purposes. First, features are selected from the whole training data set. After this, a model is trained using all the cases in the training data, but only using the selected features as predictors. Finally, once the model has been built, the model is tested against the test data and its performance is estimated using accuracy and mutual information metrics. 

\section{Feature Selection Methods}
\label{featureselectionmethods}

The aim of feature selection is to reduce the dimensionality of the structural features constructed from the raw data. Reducing the dimensionality not only reduces the computational complexity of the subsequent prediction methods, it may also lead to an improved prediction accuracy. Indeed, learning algorithms based on a smaller set of features are less prone to overfitting, i.e., the effect of erratic statistical variations of a particular observed dataset is reduced. Finally, feature selection also enhances the interpretability (visualization) of the features and understanding classifications based on them (e.g., if only two numerical features are selected, we can illustrate them by means of a scatterplot). 

Initially we also considered testing a couple of feature extraction algorithms. Feature extraction differs from feature selection in that they create new features that will be used instead of the original features. The newly created features try to maximize the variance and expressive power of the features by combining several original features into one new feature. This has a drawback that it hides the original features and makes it harder to understand the properties of the created model. E.g., in root cause analysis, it is often desirable to understand how much the outcome depends of certain features and also to understand which features have an effect to the outcome. Due to this shortcoming, we decided to not include any feature extraction algorithms into this paper.

No additional parallelization techniques were used, thus if the algorithm did not support parallelism out of the box, it was not run in parallel. The following subsections briefly describe the basics of each of the feature selection methods tested in this paper including information on the used R programming language packages and their configurations. We also briefly tested an algorithm based on \textit{Support Vector Machine} (\textit{SVM}) \cite{bennett2000support}~\cite{weston2000feature} using radial kernel, but decided to leave it out of the paper due to very poor results and extremely long response times, which were order of magnitude slower than with any of the other tested algorithms. The following subsections will briefly describe the details of all the remaining tested algorithms.

\subsection{Random Selection}
\label{method-random}

The most trivial of all the tested algorithms was a randomized selection where the desired number of features were just randomly selected from all the available features. This method was used as a baseline in order to gain a better understanding on the quality of other used selection algorithms when compared with an algorithm that does not in any way take any properties of the selected features themselves into account. This serves as a baseline selection algorithm. There should not be any algorithm that performs consistently worse than this. However, in order to alleviate the effect of inherently noisy random selections, median of three separate test runs was used in the experiments. Thus, only the test which yielded the median prediction accuracy was used as the actual result. In the graphs and analysis below, this algorithm is labeled as \textit{Random}.

\subsection{Fisher Scoring}
\label{method-fisherscoring}

Fisher scoring is based on measuring a Fisher score~\cite{DBLP:books/lib/DudaHS01} for each available feature after which N features that produced the highest Fisher score are used as the selected features. Fisher scores behave in such a way that in order for a feature to have a high value, it must have very similar values within one classification value but very dissimilar values between different classification values. In the sections below, this algorithm is labeled as \textit{Fisher}. 

%

\subsection{Feature Clustering}
\label{method-clustering}

This method is influenced by the idea provided by Covoes et al.~\cite{covoes2009cluster}. In the algorithm developed for this paper, the training data is first clustered so that every structural feature in the training set constitutes one clustering data point. Each activity sequence in the training data represents one dimension for clustering data points with values equaling the number of times that structural feature occurs within that activity sequence. K-means algorithm is then used to generate K clusters using kmeans R-function which is based on algorithm by Hartigan et al.~\cite{hartigan1979algorithm}. For each K clusters, the feature having the minimum distance to the mean of that cluster will be selected as the representative for all the features in that cluster. It should be noted also that as a side product of applying this method for feature selection, every selected feature will actually represent all the features within the same cluster. Thus, for every original feature, you have one cluster it belongs into and exactly one feature that is representing that feature in that cluster. This could be useful, e.g., in some root cause analysis scenarios.

It should be noted that K-means feature clustering, being an unsupervised learning method, does not take outcomes into account in any way and thus divides the data point space evenly without any kind of weighting or prioritization. However, clustering method, as any other feature selection method used in our paper, can easily be used in combination with other feature selection algorithms. This makes it possible to implement \textit{hybrid feature selection methods} where more than one method is used to select the final set of selected features. We used this technique to combine clustering method with several other methods to combine the properties of methods.

Four different versions of this algorithm that are covered in this paper are briefly discussed in the following subsections.

\subsubsection{Clustering Only}
\label{method-clustering-only}

For this paper we explored two different approaches One that first removed all the features having exactly the same occurrence pattern within all the cases thus removing duplicate vectors before the actual clustering step. The other variation of this algorithm does not perform this preprocessing step. 
The results of different variations being nearly the same except for the processing time, which was clearly faster with the algorithm that first dropped out all the features having exactly the same values for all the cases in the training set. Thus, we decided to limit our tests only to this algorithm variant. In the graphs shown below, this algorithm is labeled as \textit{Cluster}.

\subsubsection{Clustering with Variable Importance}
\label{method-clustering-importance}

In variable importance based feature selection, some Machine learning algorithm capable of building variable importance information, such as random forest~\cite{liaw2002classification}, is first performed. After this, the results of the algorithm are used to pick N variables having the greatest effect on the outcome. These N variables are then used as the selected features. Since the performance of variable importance algorithm itself was found to be very poor when using predictor types having hundreds of features we decided to use a hybrid approach where we first use the clustering approach described above to remove about 75\% of all the features, after which variable importance is calculated for each feature using random forest algorithm and from there, the desired number of the most important target features is picked. In this paper, randomForest -R package is first used to generate a model after which varImp -R function in Caret-package is used to extract the most important features based on the information gathered by the random forest algorithm. This algorithm is labeled as \textit{ClustImportance} in the graphs and analysis below.

\subsubsection{Clustering with Fisher Scoring}
\label{method-clustering-fisher}

This is a hybrid feature selection method where the normal clustering method described in Section~\ref{method-clustering} is performed in a way that the amount of features is brought down from the original full set of features to twice the number of features that are to be selected. If there are less than twice the number of features to pick in the first place, then this clustering step is skipped. After clustering has been performed, Fisher method is used to pick exactly the desired number of features as the final set of selected features. This method is referred to as \textit{ClustFisher} in the sections below.



\subsection{Minimum Redundancy Maximum Relevancy}
\label{method-mrmr}

This is a mutual information based approach~\cite{ding2005minimum}, which uses mutual information as a proxy for computing relevance and redundancy among features. The implementation used in this paper was provided by mRMRe -R package which claims to provide a highly efficient implementation of the mRMR feature selection via parallelization and lazy evaluation of mutual information matrix. We used ensemble method both with solution\_count set to 1, which provides results resembling classic mRMR, and also with 5, which does 5 separate runs and combines the results in the end. This time the results were also otherwise quite the same, except the 5-run version provided clearly better mutual information scores. Thus, in the graphs and analysis below, we use only 5 run version labeled as \textit{mRMREns5}.

After finding out that both mRMR and Cluster seemed to be quite efficient methods for feature selection, it was decided to include also a hybrid of these two methods in a similar fashion to ClustFisher. In this case, Clustering is used to select twice the number of desired features, after which mRMREns5 is used to make the final selection out of them. This hybrid feature selection is referred to as \textit{ClustmRMR} in the sections below.

\subsection{Least Absolute Shrinkage and Selection Operator (LASSO)}
\label{method-lasso}

This is a regression analysis method that can be used for feature selection~\cite{tibshirani1996regression}. It is related to least squares regression where the solution minimizes the sum of the squares of the errors made. The unique property for this regression technique is the usage of additional regularization that enables discarding irrelevant features and forces usage of simpler models that do not include them. Since the LASSO implementation in \textit{glmnet} R-package in itself did not provide means of sorting features by their importance and since it was not possible to directly adjust the desired number of target features, the actual used algorithm first performed 10 iterations of LASSO algorithm each yielding slightly different results. After this, all the results were collected into a single list with each feature weighted by the number of occurrences of that feature within all the LASSO results. Finally, this list was sorted from the largest height to smallest and the desired number of features were picked from the beginning of this sorted list. Two different variations of this algorithm were tested: one using lambda.1se as the prediction penalty parameter and the other using lambda.min. Due to the results being almost identical in both the cases, we selected the one using lambda.1se prediction penalty parameter. In the results below, this algorithm is labeled as \textit{LASSO1se}.

\subsection{Markov Blanket}
\label{method-markovblanket}

Markov blanket of a variable X is a minimal variable subset conditioned on which all other variables are probabilistically independent of X.~\cite{zeng2009classification} For Bayesian networks, Markov blanket of X consists of the union of the following three types of neighbors: the direct parents of X, the direct successors of X, and all direct parents of X's direct successors. Bayesian networks can be inferred from the training data after which Markov blanket for the created network is calculated by selecting the outcome as X. The result is the set of features to select. bnlearn -R package was used to perform Markov Blanket based feature selection. \textit{Hill-Climbing} algorithm is first used to construct a Bayesian network structure out of the training data. After this, Markov Blanket is extracted out of the network for the outcome feature. Finally, out of these results, the desired number of features are selected from the beginning of the returned list, or if the result does not have all the required features, only the returned features are selected.
In the results below, this algorithm is labeled as \textit{Blanket}

\subsection{Recursive Feature Elimination}
\label{method-forwardselection}

Recursive feature elimination~\cite{granitto2006recursive} starts with estimating the variable importances of all the features in the training data as in the Variable Importance -technique. After this, a smaller subset of the most important features is selected and variable importances are estimated again. This is repeated until the desired feature subset size is reached after which the resulting features can be picked. In this paper, three different variations of this method were tested: a test with only one iteration, another with two iterations and the third one with four iterations. Caret R-package's \textit{rfe} algorithm was used for these tests, with the default method based on random forests. After the initial tests, it was found out that while the accuracy and mutual information of all the cases were very close to each other, the average processing time of the 2-step algorithm was clearly better than the others. Thus, in the graphs and analysis below, we only concentrate on this algorithm labeled as \textit{Rec2S}.

\section{Test Setup}
\label{testsetup}

Testing was performed on a single system using Microsoft R Open version 3.3, Windows 10 operating system. The used hardware consisted of 3.5 GHz Intel Core i5-6600K CPU with 8 GB of memory. Tests were performed using five publicly available data sets. All the required structural features were extracted from actual event logs using the query interface of QPR ProcessAnalyzer~\cite{qprprocessanalyzer}.

Initial tests were performed on 4000 and 40000 case sample of BPI Challenge 2014 dataset~\cite{https://doi.org/10.4121/uuid:c3e5d162-0cfd-4bb0-bd82-af5268819c35}. These tests were used to make the number of tested feature selection methods smaller by dropping all the methods that clearly do not fit the interactive performance requirements that were part of the requirements. After this, all the rest of the datasets were used. Table~\ref{table:testdatasets} shows additional details of each dataset including exact numbers of extracted structural features of different types and the number of cases that belonged into the classification being predicted, which is shown in "\# Selected" -column. The table has two rows BPIC14 case, one for 4000 case sample and one for 40000 case samples. Also for both of these version, two numbers are shown in the "\# Selected" -column. The former is the number of cases having long durations and the latter the number of cases that represent a "request for information". The total number of cases in BPIC14, without any sampling, is 46616 cases.

\begin{table*}[!t]
	\centering
	\begin{tabular}{lccccccc}
		\hline
		Event Log & \# Cases & \# Selected & \# Act & \# SF & \# 2-Gram & \# Order & \# Total \\ \hline
		BPIC14-4k~\cite{https://doi.org/10.4121/uuid:c3e5d162-0cfd-4bb0-bd82-af5268819c35} & 4000 & 1441 / 581 & 39 & 20 & 772 & 1033 & 1864 \\
		BPIC14-40k~\cite{https://doi.org/10.4121/uuid:c3e5d162-0cfd-4bb0-bd82-af5268819c35} & 40000 & 8108 / 7473  & 39 & 20 & 772 & 1033 & 1864 \\
		BPIC12~\cite{https://doi.org/10.4121/uuid:3926db30-f712-4394-aebc-75976070e91f} & 13087 & 3330 & 36 & 21 & 161 & 866 & 1084 \\
		BPIC13, incidents~\cite{https://doi.org/10.4121/uuid:500573e6-accc-4b0c-9576-aa5468b10cee} & 7554 & 1579 & 12 & 8 & 75 & 129 & 224 \\
		BPIC17~\cite{https://doi.org/10.4121/uuid:5f3067df-f10b-45da-b98b-86ae4c7a310b} & 31509 & 11584 & 26 & 14 & 165 & 459 & 664 \\
		Hospital~\cite{https://doi.org/10.4121/uuid:d9769f3d-0ab0-4fb8-803b-0d1120ffcf54} & 1143 & 372 & 624 & 36 & 4272 & 79571 & 84503 \\
		\\
	\end{tabular}
	\caption{Used Event logs and numbers of features by predictor types}
	\label{table:testdatasets}
\end{table*}

All the test runs were performed using an R function that ran all the desired test runs in sequence. At the beginning of every test run, random seed was initialized. Thus, the random case samples, used in BPIC14 data set, and other random values generated within the used algorithms behaved the same way in every run, provided that the algorithm used random -methods that support setting the seed using \textit{set.seed} -R function. 

In the tests, the training data was first extracted so that 25\% of the provided data rows were randomly selected. This training data was first used by the feature selection algorithm to be tested, after which it was used to build the classification \textit{GBM} model for predicting given phenomenon. This classification model was then used for measuring the performance of the feature selection. Mutual information metrics were approximated also at this final phase.

The first run of tests was performed using test data having 4000 cases extracted from the full BPI Challenge 2014 data set. For this first run, all the algorithms were tested so that the number of selected features were 10 and 30. For each of these combinations, 13 different sets of feature patterns were selected. The selected structural feature patterns were different combinations of the following patterns described in Table~\ref{table:examplestructuralfeatures}: activity, starter and finisher, 2-grams and ordering. In BPIC14-event logs, activity and 2-grams features included occurrence counts, while all the other feature types were just boolean values indicating whether the feature occurs at least once in a case. In all the other datasets, all the features are occurrence counts. 

The combinations of used feature types were created in a way that all the possible combinations of the patterns were tested where activity pattern was present. We also included some other more interesting combinations, thus generating 11 different pattern combinations. In addition to these, we also tested 2-grams and ordering separately, as well as having both 2-grams and ordering. As discussed in the previous sections, we did not want to include any other patterns due to the number of potential new features that would have been needed in order to cover all the possible cases. E.g., adding 3-grams would potentially have generated $N^{3}$ additional features where N equals the number of different activities in the training data, which in this case is 39 yielding the maximum of 60000 new features.

All the tests performed on the first data set were run to predict two different outcomes, which are later in this paper referred to as scenarios. The first scenario was whether the case duration is longer than 7 days. In this case, nearly 36\% of all the cases in the small test set had this outcome. This is an example of a prediction that can be trained directly from the event information without any need for additional case or event attributes. The second scenario that was tested is based on additional case-level information provided with the event data: Does the case represent a "request for information" or something else such as an "incident"? In this case, nearly 15\% of all the test cases in the small sample had this outcome. For all the tests performed on the same sample size, the actual used cases and their predictors were always the same.

For all the other data sets, a classification was made based on the duration of cases. In BPIC12 and BPIC13, duration threshold was set to 2 weeks. In BPIC17, 4 weeks was used as threshold. In Hospital data set, 20 weeks was used.

\section{Test Results}
\label{testresults}

\begin{figure}[!t]
	\begin{minipage}{\linewidth}
		\begin{figure}[H]
			\centering
			\includegraphics[trim={1.9cm 10.3cm 2cm 11.5cm},clip,width=\linewidth]{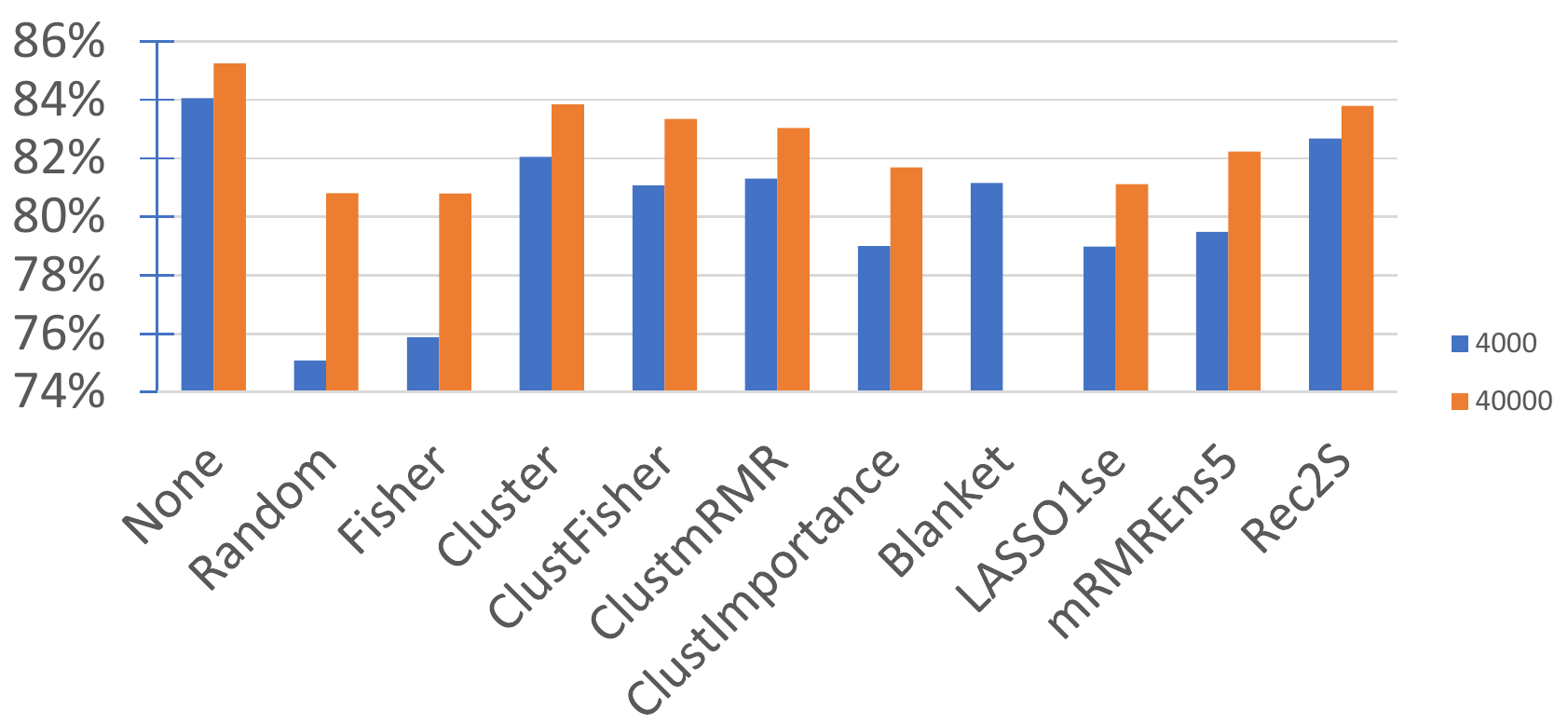}
			\caption[]{Average accuracy of all the tested algorithms.}
			\label{figure:allalgorithmsacc}
		\end{figure}
	\end{minipage}
	\begin{minipage}{\linewidth}
		\begin{figure}[H]
			\centering
			\includegraphics[trim={1.9cm 11.1cm 2cm 11.6cm},clip,width=\linewidth]{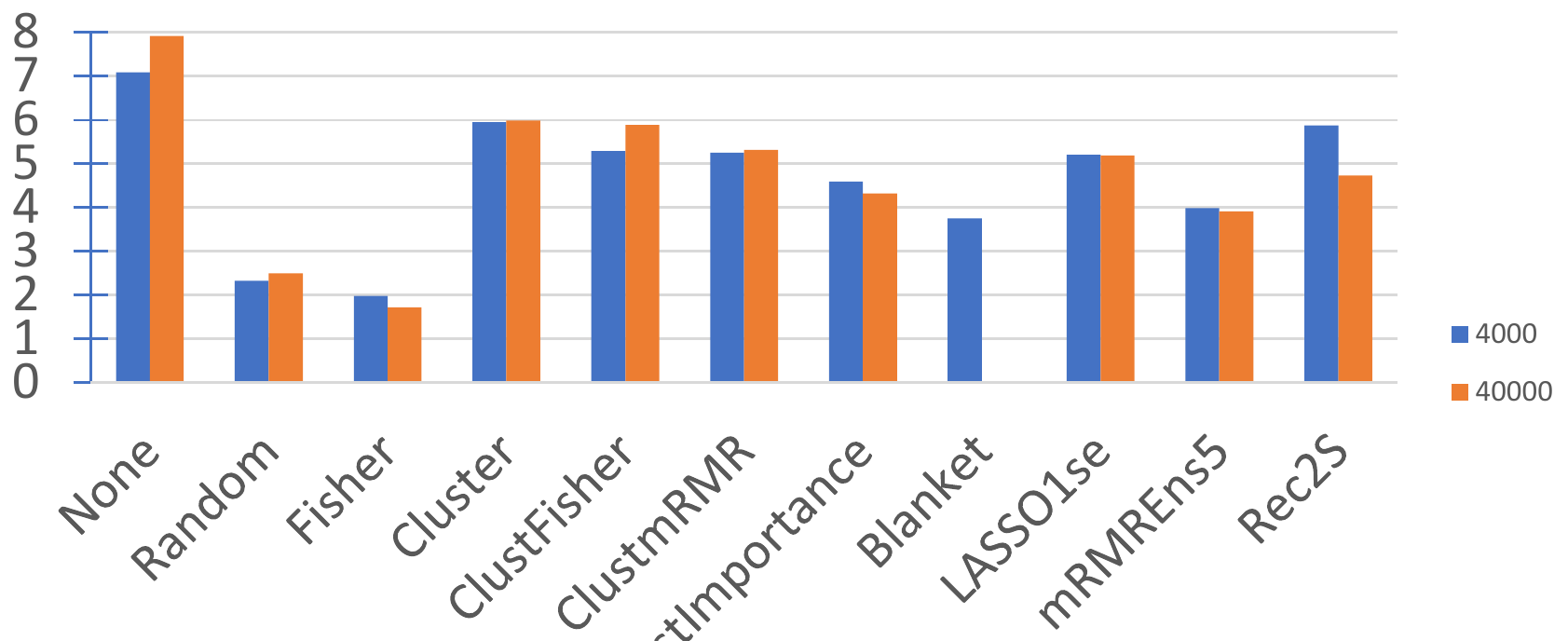}
			\caption[]{Average mutual information of all the tested algorithms.}
			\label{figure:allalgorithmsmut}
		\end{figure}
	\end{minipage}
\end{figure}

We began the actual analysis of our first round of tests by estimating the average classification accuracy of all the tested algorithms for all the tested feature counts and all the tested structural feature patterns using both the test scenarios in 4000 case sample. After the first test run, a second run was made with the same BPIC14 data set, but this time with a sample of 40000 cases. At this point we also decided to not include Blanket method into this test run due to its very slow performance and often did not manage to finish at all. The results of these test runs are shown in Figure~\ref{figure:allalgorithmsacc}. The first column on the chart labeled \textit{None} shows the accuracy achieved by not performing any feature selection. 
Each column in the chart represent an average of 44 test run results, except for \textit{None} column which represents the average of 22 test run results, since selecting different number of features is not applicable for it. Based on these results, we can immediately see that increasing the sample and also training set size increases also classification accuracy. We can also see that for 4000 case sample, the top three feature selection algorithms ordered in the descending order of accuracy are: Recursive, Cluster and Cluster with mRMR. However, when the sample size is increased to 40000 cases, the order is changed to: Cluster, Recursive, Cluster with Fisher while mRMR dropped to 5th place.

With the same test data, we also measured the average mutual information score of the data set filtered using feature selection algorithm. This is shown in Figure~\ref{figure:allalgorithmsmut}. In this chart, None shows the absolute maximum score achievable by any feature selection algorithm. The top three ranking of algorithms when ranked by mutual information is: Recursive, Cluster and mRMR. With 40000 case dataset, the ranking order is Cluster, Cluster with Fisher and LASSO. When the mutual information is calculated between the result of the feature selection algorithm and the expected outcome, the ranking is for smaller sample: Recursive, Cluster and Cluster with mRMR. For bigger sample this becomes: Cluster, Cluster with Fisher and LASSO, while mRMR can be found at 7th place. 

After this, we analyzed the response times for all the tested feature selection algorithms with the same test data. This time however, we did not include starter and finisher predictor types since they would have made the readability of the figure much worse and also would not have provided much additional information due to the small effect they have into the results in the tested scenarios.

\begin{figure}[!t]
	\begin{minipage}{\linewidth}
		\centering
		\includegraphics[trim={2cm 10.7cm 2cm 12.1cm},clip,width=0.8\linewidth]{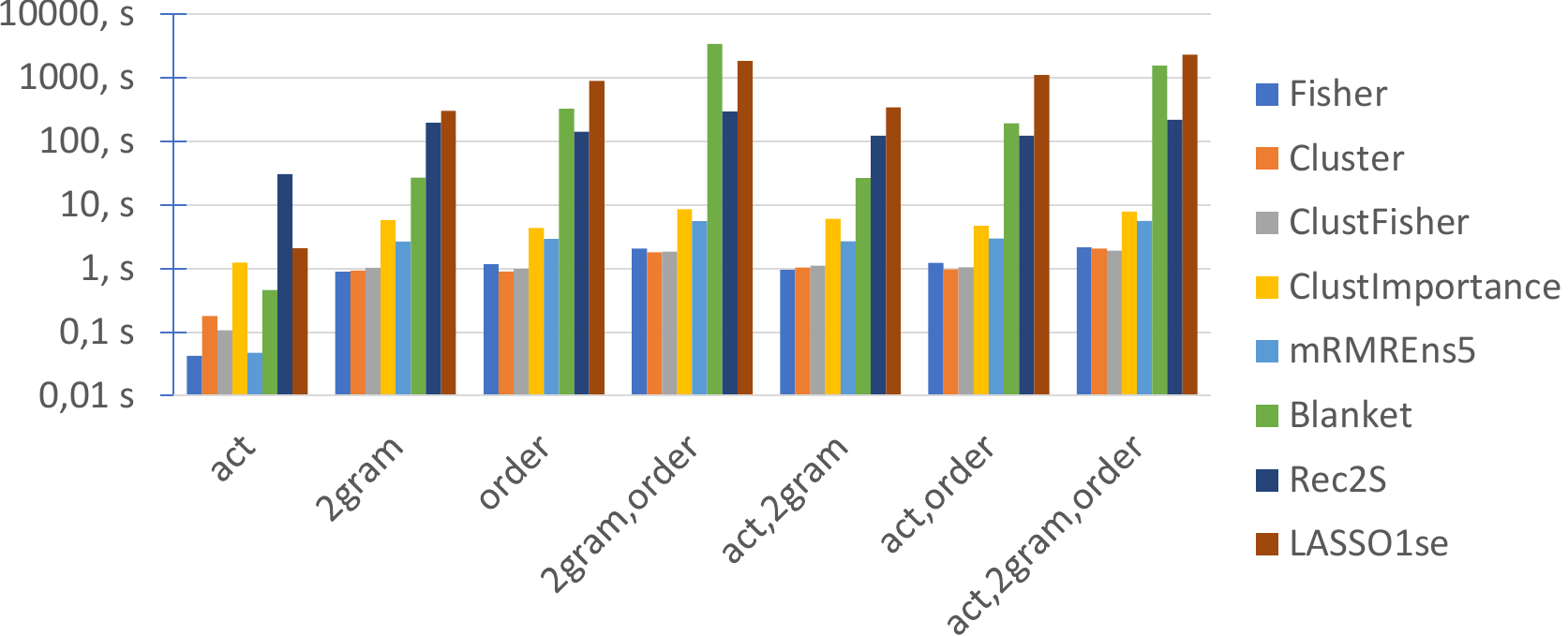}
		\caption{Average feature selection response time of all the tested algorithms.}
		\label{figure:allalgorithmsproctime}
	\end{minipage}
\end{figure}

As seen in the Figure~\ref{figure:allalgorithmsproctime}, the time required to perform feature selection for the tested algorithms and predictor types varied very much. Each column in this chart represents an average time required by 4 test runs. In the worst cases, the difference between the slowest and the fastest algorithm was three orders of magnitudes, with Fisher, Cluster, Cluster hybrids and mRMR usually performing much faster than all the rest. Out of these, Clusters usually slightly outperformed mRMR. With the 40000 case dataset, the most notable change is that mRMR now comes very close to, and sometimes even outperforms, Fisher while both of them being almost one order of magnitude faster than Cluster algorithms which in turn were at least one order of magnitude faster than the LASSO and Recursive algorithms. Surely better performance could have also been achieved by optimizing the used R -scripts or by using some tailor-built natively compiled software, especially for algorithms that had more scripted parts such as Fisher and Cluster algorithms.

The largest performance variations within one algorithm were measured using Blanket algorithm which performed in the fastest predictor types, almost as fast as the fastest algorithms, but in the slowest predictor types, it performed almost over four orders of magnitudes slower. 

Based on this data, it was decided to drop Blanket, LASSO1se and Rec2s from any further examinations. We also decided to drop Fisher and Cluster Importance due to their somewhat poor classification accuracy compared to other remaining algorithms.

\begin{figure}[!t]
	\begin{minipage}{\linewidth}
		\begin{figure}[H]
			\centering
			\includegraphics[trim={2cm 11.5cm 2cm 12.8cm},clip,width=0.8\linewidth]{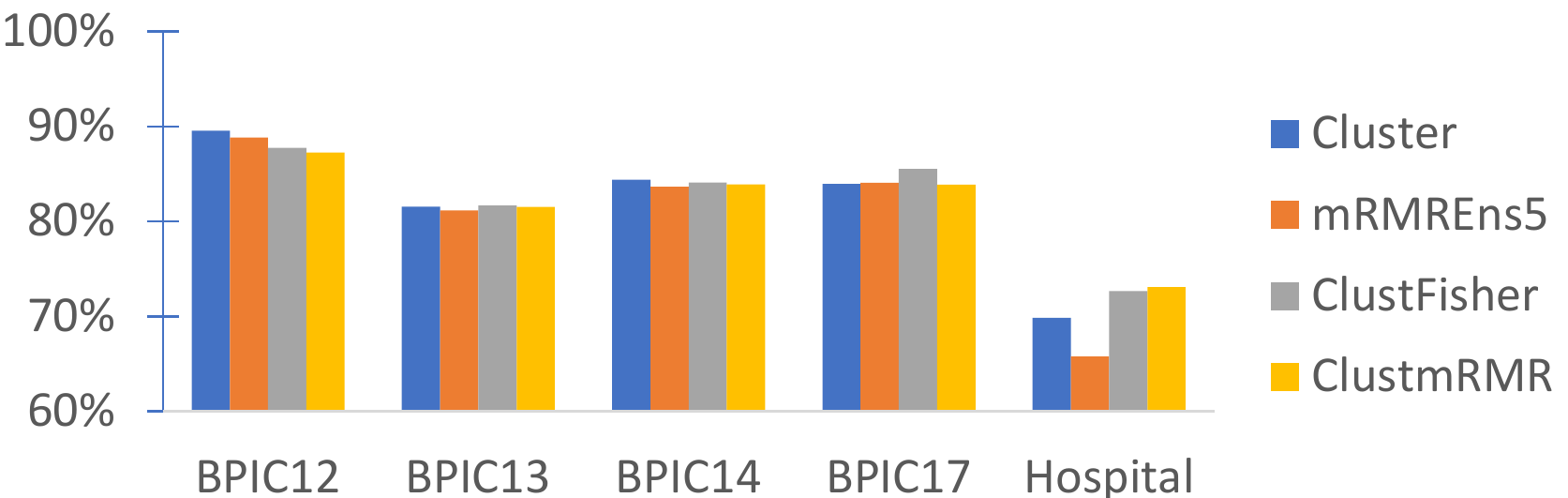}
			\caption{Average accuracy of the algorithms using all the available predictor types.}
			\label{figure:alldatasetsallpredsets}
		\end{figure}
	\end{minipage}
	\begin{minipage}{\linewidth}
		\begin{figure}[H]
			\centering
			\includegraphics[trim={2cm 11.5cm 2cm 12.8cm},clip,width=0.8\linewidth]{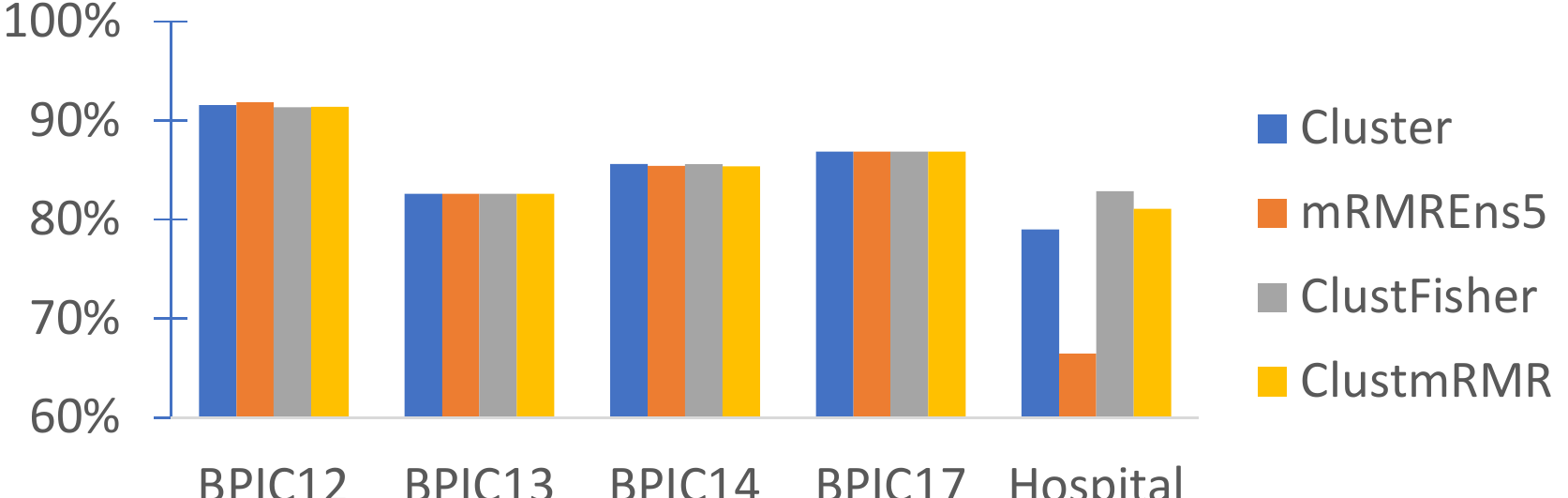}
			\caption{The maximum accuracy of the algorithms using all the available predictor types.}
			\label{figure:alldatasetsallpredsetsmax}
		\end{figure}
	\end{minipage}
\end{figure}

\begin{figure}[!t]
	\begin{minipage}{\linewidth}
		\begin{figure}[H]
			\centering
			\includegraphics[trim={2cm 11.5cm 2cm 12.9cm},clip,width=0.8\linewidth]{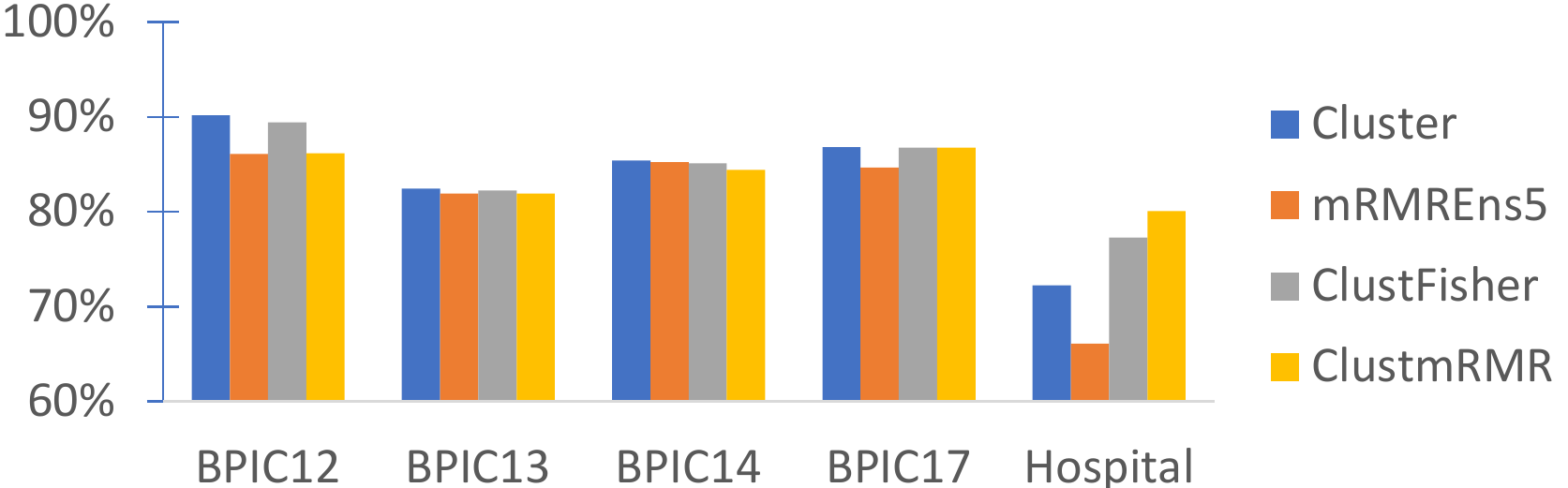}
			\caption{Average accuracy of the algorithms using only \textit{activity} predictor type.}
			\label{figure:alldatasetsact}
		\end{figure}
	\end{minipage}
	\begin{minipage}{\linewidth}
		\begin{figure}[H]
			\centering
			\includegraphics[trim={2cm 11.5cm 2cm 12.9cm},clip,width=0.8\linewidth]{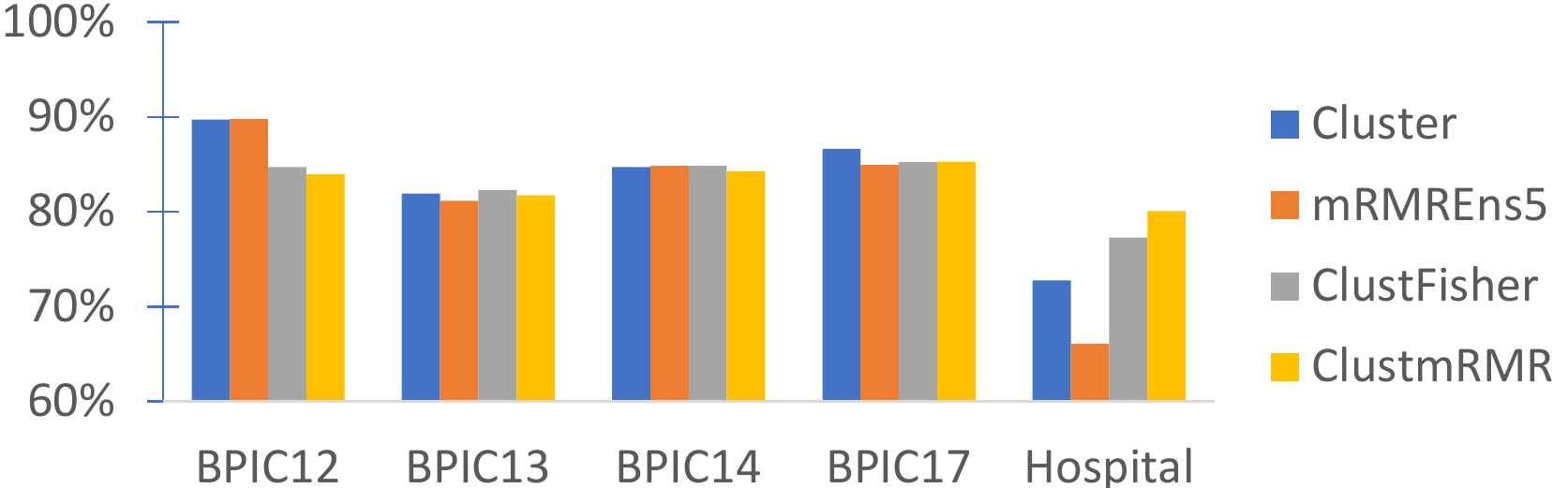}
			\caption{Average accuracy of the algorithms using both \textit{activity} and \textit{2-gram} predictor types.}
			\label{figure:alldatasets2gram}
		\end{figure}
	\end{minipage}
	\begin{minipage}{\linewidth}
	\begin{figure}[H]
		\centering
		\includegraphics[trim={2cm 12.5cm 2cm 13.8cm},clip,width=\linewidth]{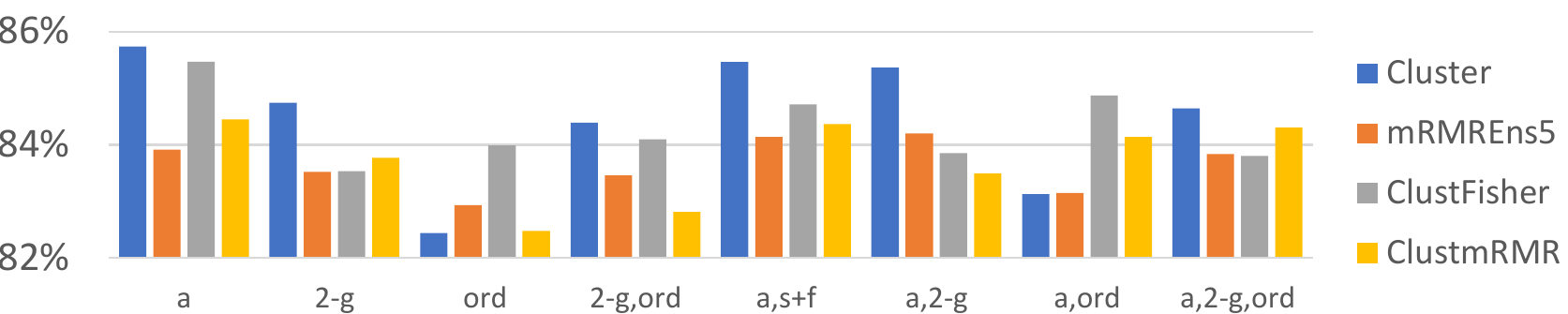}
		\caption{Average accuracy by algorithm and by predictor type for all the tested BPIC event logs.}
		\label{figure:accdetails}
	\end{figure}
\end{minipage}
\end{figure}

Next we used all the remaining algorithms to run all the tested test parameter combinations on several other datasets. Figure~\ref{figure:alldatasetsallpredsets} shows the accuracy achieved when using all the tested predictor types as source features for the feature selection using the narrowed down set of algorithms and all the tested datasets. mRMREns5 columns are missing in Hospital-data set in the figure because the used R implementation of mRMR algorithm supported only maximum of 46340 features that was exceeded in that case.
Figure~\ref{figure:alldatasetsact} shows otherwise the same information, except this time only structural features of \textit{activity} predictor type are included as source features. Figure~\ref{figure:alldatasets2gram} shows the same information when only features of \textit{2-gram} predictor type are available. Finally, Figure~\ref{figure:accdetails} shows how the accuracy of the tested algorithms change when the available predictor types are changed.


Based on these figures, it can be seen that mRMR and Clustering both provide nearly as accurate results. BPIC17 is the only tested dataset where mRMR provides slightly better average accuracy than Cluster over all the tested predictor types. However, also in this case, mixing both Cluster and Fisher provided even better accuracy. Hospital dataset, which has very few cases and lots of different types of structural features, was the only dataset where hybrid cluster algorithms provided clearly better results than the normal Cluster dataset. It also seems that when the amount of cases is low compared to the amount of features, mRMR does not manage to get as accurate results as the Cluster algorithms. Generally, based on all these tests, it can be said that the order from the most accurate to the least accurate algorithm is: Cluster, Cluster Fisher, mRMR and Cluster mRMR. However, the difference between the worst and the best, based on 72 experiment results for every algorithm is only 0,7~\%, which is not much. One small detail to note is that despite mRMR getting slightly inferior accuracies, it managed to get the best accuracy classification of all the experiments for BPIC12 dataset using only \textit{2-gram} predictors where Cluster managed to get its slightly worse value using \textit{activity}, \textit{starter} and \textit{finisher} predictors.  It seems that having hybrid algorithm only helps when the amount of available features is very high and the percentage of available relevant features is low, as is the case in the Hospital dataset. This clearly indicates that Clustering performs the clustering without taking the outcome into account at all, while hybrid version manage to get better results since they finalize the feature selection by trying to select those clusters that have the greatest impact to the outcome. One interesting finding from Figure~\ref{figure:accdetails} is also that the accuracy of mRMR can be improved from the accuracy achieved by using only \textit{activity} predictors by using also either \textit{2-gram} predictors or both \textit{starter} and \textit{finisher} predictors. This does not work for Clustering at least in BPIC datasets, since it always achieves the best average accuracy by using only \textit{activity} predictors.

Since Clustering, Clustering hybrids and mRMR performed so well in this analysis that they could be incorporated without changes into some interactive process mining systems preferring under ten second response times when the size of the event log used for training is close to 1000 cases, we took Clustering and mRMR for closer inspection now emphasizing especially on the performance of different predictor types.

\begin{figure}[!t]
	\begin{minipage}{\linewidth}
		\centering
		\begin{minipage}{0.45\linewidth}
			\begin{figure}[H]
				\includegraphics[trim={1.9cm 8.1cm 2cm 9.2cm},clip,width=\linewidth]{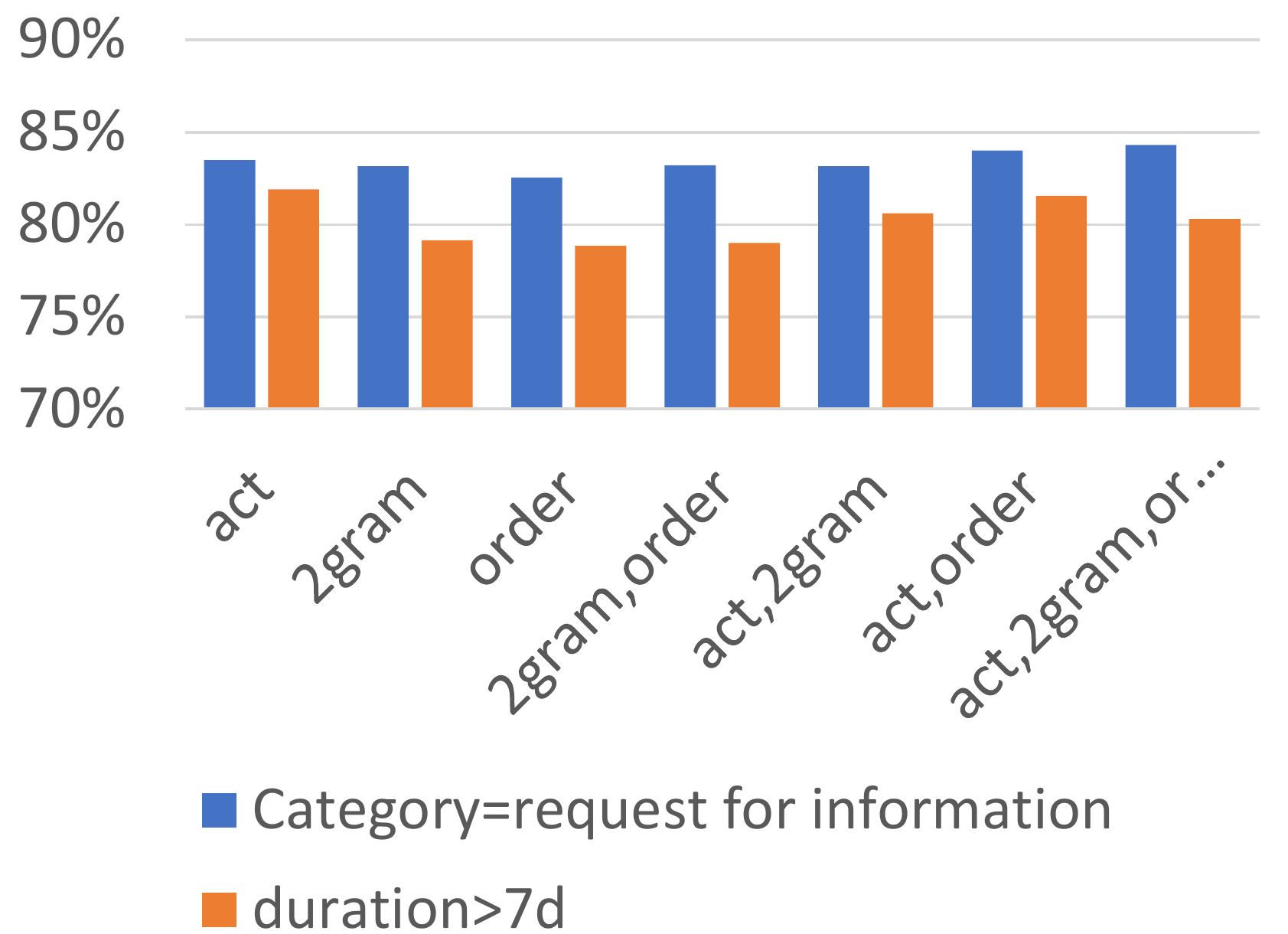}
				\caption{Average accuracy for Cluster algorithm by predictor types  separately for each scenario for 4k case sample in BPIC14.}
				\label{figure:clusteraccdetails}
			\end{figure}
		\end{minipage}
		\hspace{0.05\linewidth}
		\begin{minipage}{0.45\linewidth}
			\begin{figure}[H]
				\includegraphics[trim={1.9cm 8.1cm 2cm 9.2cm},clip,width=\linewidth]{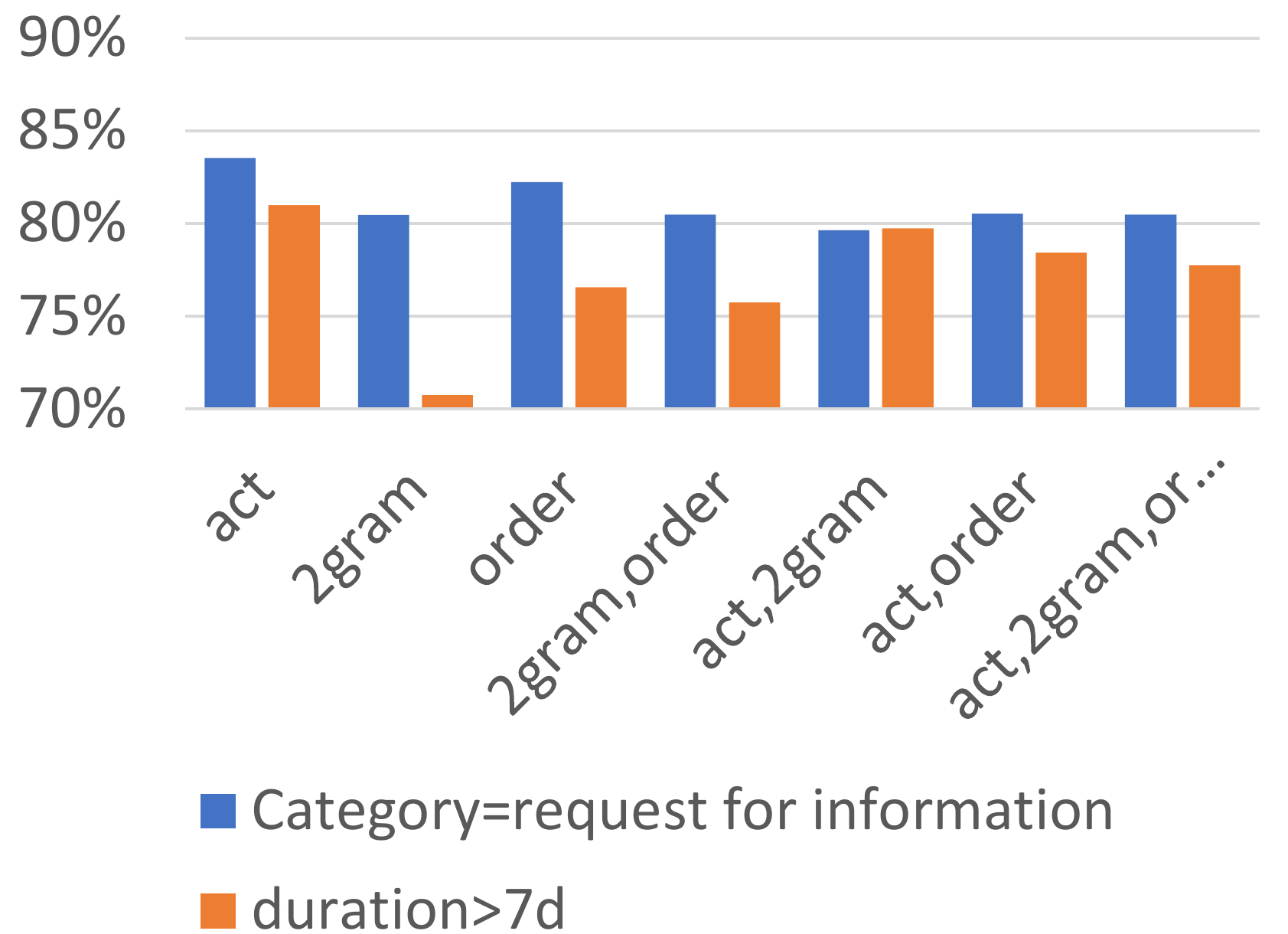}
				\caption{Average accuracy for mRMR algorithm by predictor types  separately for each scenario for 4k case sample in BPIC14.}
				\label{figure:mrmraccdetails}
			\end{figure}
		\end{minipage}
	\end{minipage}
\end{figure}

First, we analyzed the classification accuracy of both the algorithms separately for both the tested scenarios in BPIC14 event log having only 4000 cases. Figure~\ref{figure:clusteraccdetails} shows this information for Cluster algorithm and Figure~\ref{figure:mrmraccdetails} for mRMR algorithm. From these results, it can be seen that there is a lot of variation between the two tested scenarios in the BPIC14 dataset. In both the cases and in almost all the predictor types, predicting case duration produced clearly worse results than in the categorization scenario. It seems that mRMR algorithm was not able to get any additional accuracy into its predictions by including any additional predictor types on top of \textit{activity} predictors, whereas Cluster algorithm managed in the categorization case to get better accuracy when adding \textit{order} type predictors in addition to \textit{activities}. 

It should be also noted that the time required for building a classification model with a feature selection algorithm selecting 10 features was only about 1\% - 3\% of the total time required when building the model with all the 1864 tested structural features without any feature selection. When using 3000 cases to build a model, the total measured time difference in the test system was about 250 seconds. During this time, it would have been possible to run the clustering feature selection several times. Thus, it is clear that having a feature selection performed before model building, at least when GBM is used, is essential when trying to improve the time required for model building.

As additional notes about the Dutch academic hospital dataset~\cite{https://doi.org/10.4121/uuid:d9769f3d-0ab0-4fb8-803b-0d1120ffcf54}, the tests performed in this dataset indicated that it is absolutely critical to perform some kind of feature selection before training the model since building the model without any selection failed when attempting to use all the 79571 ordering features valid in this event log. The amount of features was actually so huge that some of the tested algorithms failed completely when trying to select the relevant features. 

Thus, in order to fulfill all the requirements, we have for the algorithm, based on the performed tests, the best option from the selected set of algorithms is Cluster based feature selection with only \textit{activity} type predictors with values being the occurrence counts of \textit{activities} within a \textit{case}. In our experiments, it gave the best overall trade-off in performance considering mutual information, classification accuracy and response time. For best possible response times, mRMR might be a better alternative with only \textit{activity} predictors. Cluster Fisher is the recommended hybrid selection algorithm that performs in average slightly worse than plain Cluster when only \textit{activity} type predictors are used, but when more features are available, it seems to perform slightly better.



\section{Related Work}
\label{relatedwork}

In addition to the related work related to more complex structural features referred already in Section\ref{complexstructuralfeatures}, several papers have been written on the subject of applying data mining and machine learning techniques into predicting outcomes of the business processes. In \cite{DBLP:conf/cidm/BoseA13}, the authors present a framework that is capable of automatically detecting "signatures" that can be used to discriminate between desired and undesired behavior within traces both seen or unseen. These signatures are essentially combinations of structural features similar to those described in Section~\ref{structuralfeatures}. This paper does not in itself specify any automatic feature selection method. Instead, the user is required to specify manually the desired activity sequence patterns, referred to as sequence feature types. After this all the matching features will be used for signature detection. Thus, our research complements the research made in this paper by experimenting with different automatic feature selection methods that could be applied before this signature detection phase in order to reduce the computational cost of signature detection at the cost of some prediction accuracy.

In \cite{DBLP:conf/otm/NguyenDRMS14}, the authors evaluate the accuracy achieved with three different classification methods using several combinations of more complex structural feature patterns discussed in Section~\ref{complexstructuralfeatures} for three different datasets. As result, they find out that just having Activity frequencies often yield, if not the best, then at least almost as good results as the best tested structural feature pattern combination. This finding is visible also in our tests as shown in Figures~\ref{figure:clusteraccdetails} and \ref{figure:mrmraccdetails}.

In \cite{DBLP:journals/corr/Francescomarino15}, the authors present a framework for predicting outcomes of user specified predicates for running cases using clustering based on activity sequence prefixes and classification using attributes associated to events. In \cite{DBLP:conf/bpm/LeontjevaCFDM15}, the authors have assessed the benefits of including case and event attributes when performing predictions based on sequences of activities. In \cite{DBLP:conf/bpm/TeinemaaDMF16}, the authors present a predictive process monitoring framework that is also able to mine unstructured textual information embedded into attributes related to events. In \cite{DBLP:journals/dss/ConfortiLRAH15}, the authors propose a recommendation system that automatically determines the risk that a fault will occur if the input the user is giving to the system will be used to carry on a process instance.

Until now there has not been systematic testing of applying automatic feature selection algorithms after selecting structural feature patterns to use and before building models used for classification. The aim of this feature selection is to minimize the computational cost of the building of classification models. Creating such an approach is crucial for obtaining predictions with interactive response time requirements. This is the primary contribution of this paper.

\section{Conclusions}
\label{conclusions}

In this paper, we have designed a system for assessing the performance and response times of selected feature selection algorithms specifically tuned into the context of selecting structural features extracted from properties of sequences of activities derived from event logs. Using this system, we tested nine feature selection techniques. 

Each algorithm was tested first using a publicly available real-life Rabobank Group ICT dataset and tuned for two different classification use cases: Predicting whether the duration of a case is longer than seven days, and classifying whether a case is of type request for information. Most of the tests were also run using two different sample sizes out of the full dataset. For sanity checking and benchmarking purposes, we also added test runs without any feature selection and also with randomized feature selection. Finally, for a smaller sub-set of algorithms, we ran additional duration based classification tests on four other publicly available data sets.

We also proposed a rough categorization method for some of the types of structural features that can be extracted from event logs. In this paper, we selected four of these types for closer inspection.

As summary for all the tests and their results, it can be clearly seen that structural features can provide additional means for improving the precision to classifications made for cases in event logs. When the number of selected features is small, the most efficient source of features is activities. Increasing the number of features improves the classification accuracy, but also while doing so, best results are achieved by adding features from other structural feature types such as event type orderings into the set of structural features from which the feature selection is made. However, there is a drawback that having a bigger pool of features to select from makes creating classification models as well as the feature selection slower. As our goal was also to find an algorithm that could perform feature selection and classification with interactive response times using the sample sizes used in this paper, we found out that only one feature selection algorithm of the tested algorithms provided both the speed and accuracy required for the task. 

According to the tests, the most consistently well performing algorithm was \textit{Cluster} algorithm we developed for this paper which first used \textit{k-means} algorithm for clustering features into the desired number of clusters by having cases as clustering dimensions, after which the features closest to the center of each cluster were selected as the selected features. 

This algorithm performed especially well when the available structural features did not have many redundant features that did not have an effect into the final classification or when the available training data did not cover very well all the available features. In those cases, hybrid algorithms, such as one mixing both the \textit{Clustering} and \textit{Fisher} scoring, seem to outperform \textit{Cluster} algorithm. Our partially self implemented \textit{Cluster} algorithm was not, especially with larger number of cases, as fast as another quite well performing \textit{mRMR} component. Both of these algorithms lose, in average accuracy in some data sets, to \textit{Recursive Feature Selection}, but due to its decades slower response time, it can not be recommended due to the interactive usage requirements set in this paper. For computer assisted root cause analysis, in addition to providing the list of the most important features, \textit{Cluster} algorithm provides also a mapping from each of the original structural features to one selected feature that most closely resembles the original feature in the set of selected features.

All the raw information gathered from over 2300 successfully performed test runs can be found in the support materials~\cite{supportmaterials}. This raw information, some of which was not discussed nor explored in this paper in detail include also: Measured accuracies, mutual information scores, computation times, selected features and confusion matrices.

\section{Acknowledgements}
\label{acknowledgements}

We want to thank QPR Software Plc for funding our research. Financial support of Academy of Finland projects 139402 and 277522 is acknowledged.

\label{references}
\bibliography{paper}

\end{document}